\documentclass[a4paper,twoside]{article}
\usepackage{epsfig}
\usepackage{subfigure}
\usepackage{calc}
\usepackage{amssymb}
\usepackage{amstext}
\usepackage{amsmath}
\usepackage{amsthm}
\usepackage{multicol}
\usepackage{pslatex}
\usepackage{apalike}
\usepackage{SciTePress}
\usepackage[small]{caption}
\usepackage{url}

\begin{document}

\title{On Reducing the Number of Visual Words in the Bag-of-Features Representation} 

\author{\authorname{Giuseppe Amato\sup{1}, Fabrizio Falchi\sup{1} and Claudio Gennaro\sup{1}}
\affiliation{\sup{1}Institute of Information Science and Technologies, CNR, via Moruzzi, 1, Pisa, Italy}
\email{\{f\_author, s\_author\}@isti.cnr.it}
}



\abstract{
A new class of applications based on visual search engines are emerging, especially on smart-phones that have evolved into powerful tools for processing images and videos. The state-of-the-art algorithms for large visual content recognition and content based similarity search today use the ``Bag of Features'' (BoF) or ``Bag of Words'' (BoW) approach. The idea, borrowed from text retrieval, enables the use of inverted files. A very well known issue with this approach is that the query images, as well as the stored data, are described with thousands of words. This poses obvious efficiency problems when using inverted files to perform efficient image matching.
In this paper, we propose and compare various techniques to reduce the number of words describing an image to improve efficiency and we study the effects of this reduction on effectiveness in landmark recognition and retrieval scenarios. We show that very relevant improvement in performance are achievable still preserving the advantages of the BoF base approach.}

\onecolumn \maketitle \normalsize \vfill

\section{\uppercase{Introduction}}
\label{sec:introduction}
\noindent The use of local features, as for instance
SIFT \cite{SIFT}, has obtained an increasing appreciation during the last decade, for its good performance in tasks like image matching, object recognition, landmark recognition, and image classification.
Briefly, with these techniques an image visual content is described
by identifying a set of (interest) points and by describing the
region around them with histograms (the local feature), as for
instance histograms of brightness gradients. The image match task
is executed by first matching the local features, and then by
checking if there is some (geometric) consistency between matched
pairs of interest point, to decide if images, or objects in
images, match as well.

The total number of local features extracted from an image depends on various setting of the feature extraction tools. However, typically it is of the order of some thousands. As a consequence, matching an image against a database of images becomes a very challenging task from the efficiency point of view. For instance, if the database contains one million images and on average every image has one
thousand local features, matching a query image against this database requires matching 1,000 different local features, extracted from the query, against one billion local features, extracted from the database.

In order to mitigate this problem, some years ago the Bag of
Feature approach (BoF) was proposed in \cite{Sivic2003}.
The BoF
approach quantizes local features extracted from images
representing them with the closest local feature chosen from a
fixed visual vocabulary of local features (visual words). In this
way, images are no longer represented by a set of identifiers of visual
words from the visual vocabulary that is used to replace the
original local features.
Matching of images represented with the BoF approach is performed
with traditional text retrieval techniques and by
verifying their (geometric) consistency. This process can be
executed more efficiently, than linearly scanning the entire
database, by using inverted files \cite{Salton1986} and search
algorithms on inverted files.

However, even if inverted files offer a significant improvement of
efficiency, with respect to a trivial sequential scan search
algorithm, in many cases, efficiency is not yet satisfactory. A
query image is associated with thousands of visual words.
Therefore, the search algorithm on inverted file has to access
thousands of different posting lists of the inverted file.
As mentioned in \cite{Zhang2009}, "a fundamental difference between an image query (e.g. 1500 visual terms) is largely ignored in existing index design. This difference makes the inverted list inappropriate to index images."
From the very beginning \cite{Sivic2003} some words reduction techniques were used (e.g. removing 10\% of the more frequent images). However, as far as we know, no experiments have been reported on the impact of the reduction on both efficiency and efficacy.

To improve efficiency, many different approaches have been considered including
GIST descriptos \cite{Douze2009}, Fisher Kernel \cite{Zhang2009} and Vector of Locally Aggregated Descriptors (VLAD) \cite{Jegou2010}.
However, their usage does not allow the use of traditional text search engine which has actually been another benefit of the BoF approach.

In order to mitigate the above problems, this paper proposes,
discusses, and evaluates some methods to reduce the number of
visual words assigned to images.
We will see that it is possible to
significantly reduce their number with a very minor degradation of
the accuracy, and with a significant efficiency improvement.
Specifically, we propose and discuss methods based on the use
of the \emph{scale} of the local features
that can be applied before the visual words have been assigned
and also methods based on statistics of the usage of visual words in images
(using the term frequency \emph{tf}), across the database (relying on the inverse document frequency (\emph{idf}), and
on the \emph{tf*idf} combination \cite{Salton1986}).
We also perform experiments using \emph{random} reduction as a baseline.
The \emph{tf*idf} approach was also presented  in \cite{Thomee2010} using the SURF descriptor. However, in their work the authors did not present any  comparison with other approaches.
The effectiveness of the approaches is measured on landmark retrieval and recognition tasks for which local features and in particular the BoF approach is today
considered the state-of-the-art.
Experiments were conducted on three dataset for testing both retrieval and recognition scenarios.

\section{\uppercase{Selection Criteria}}
\label{sec:criteria}
\noindent The goal of the BoF approach is to substitute each
description of the region around an interest points
(i.e., each local feature) of the images with visual words obtained
from a predefined vocabulary in order to apply traditional
text retrieval techniques to content-based image retrieval.

The first step to describe images using visual words is to select
some visual words creating a vocabulary. The visual vocabulary is
typically built grouping local descriptors of the dataset using a
clustering algorithm such as \emph{k-means}.
The second step is to
assign each local feature of the image to the identifier of the
first nearest word in the vocabulary. At the end of the process,
each image is described as a set of visual words.
The retrieval phase is then performed using text retrieval techniques considering a query image as disjunctive text-query.
Typically, the \emph{cosine} similarity measure in conjunction with a term weighting scheme is adopted for evaluating the similarity between any two images.

In this section we present five criteria for local features and
visual words reduction.
Each proposed criterion is based on the definition of a score
that allows us to assign each local feature or word, describing an
image, an estimate of its importance. Thus, local features or
words can be ordered and only the most important ones can be
retained. The percentage of information to discard is configurable
through the definition of a score threshold, allowing trade-off between efficiency and effectiveness.

The criteria we tested are:

\begin{itemize}
\item{\textbf{\emph{random}}} --
A very naive method to reduce the number of words assigned to an
image is to randomly remove a specific percentage of local
features in the image description. This method is used as a baseline
in our experiments.

\item{\textbf{\emph{scale}}} --
Most of the local features defined in the last years (e.g., SIFT
and SURF) report the scale at which the feature was extracted for
each keypoint.
The fact that extraction is not performed at the original resolution
is actually the main reason for the scale invariant.
Descriptions and interest points detected at higher scale should
be also present at lower resolution versions of the same images or
of the same object.
The intuition is that the bigger the scale the higher the importance.
This approach can be performed before the words assignment phase
increasing performance also during the words assignment, since the cost of assigning words to images is linear with the
number of local features.
Please note that the scale threshold is not defined a priori but it depends on the number of local features actually extracted from the image.

\item{\textbf{\emph{tf}}}--
During the BoF words assignment phase, each local
feature is substituted with the identifier of the nearest word in the visual vocabulary.
Thus, after this step every image is described with a set of visual words.
Typically a word appears more than once in an image description because distinct but similar local features in the original description were substituted by the very same visual word.
A possible approach for words reduction is to remove the words having the lowest number of occurrences.
In this case we are ordering the words with respect to their term frequency (\emph{tf}) in the image.
\cite{Salton1986}.

\item{\textbf{\emph{idf}}} --
When words have been assigned to all the images in the dataset, it
is possible to evaluate the inverse document frequency (\emph{idf})
\cite{Salton1986} of all the features in the vocabulary. In
Information Retrieval words with highest \emph{idf} are considered more
important than the others \cite{Salton1986}.
Note that depending on the relative \emph{idf} values of
the words describing an image, the same word could be discarded
for a given image and retained in another.

\item{\textbf{\emph{tf*idf}}} --
In information retrieval a very popular strategy to assign
relevance to words is the \emph{tf*idf} approach \cite{Salton1986}. This
strategy states that the relevance of a word in a document is
obtained by multiplying its \emph{tf} in the given document by its \emph{idf}.
We can use the same strategy to order the visual words in an image
and discard first the words with smaller \emph{tf*idf}.
\end{itemize}

\section{\uppercase{Experiments}}
\label{sec:exp}
\noindent The effectiveness of the approaches is measured on both a image retrieval and a landmark recognition tasks using two distinct datasets. Efficiency is also tested on a larger professional dataset intended for similarity search.
In the following we describe the recognition system, the performance measures, the datasets and we discuss the experimental results obtained.


The retrieval engine used in the experiments is built as following:
\begin{enumerate}
\item For each image in the dataset the SIFT local features are extracted for the identified regions around interest points.
\item A vocabulary of words is selected among all the local features using the \emph{k-means} algorithm.
\item The \emph{Random} or \emph{Scale} reduction technique is performed (if requested).
\item Each image is described following the BoF approach, i.e., with the ID of the nearest word in the vocabulary to each local feature.
\item The \emph{tf}, \emph{idf}, or \emph{tf*idf} reduction technique are performed (if requested).
\item Each image of the test set is used as a query for searching in the training set. The similarity measure adopted for comparing two images is the Cosine between the query
vector and the image vectors corresponding to the set of words assigned to the images. The weight assigned to each word of the vectors are calculated using \emph{tf*idf} measure.
\item In case the system is requested to identify the content of the image, the landmark of the most similar image in the dataset (which is labeled) is assigned to the query image.
\end{enumerate}

Typically, the result obtained with the \emph{tf*idf} weighting and \emph{cosine} similarity measure using inverted index is reordered considering geometric checks based on RANSAC (Random Sample Consensus).
However, in this paper we focus on optimizing the number of words to improve efficiency of the search performed through the inverted files and thus we do not leverage on geometric consistency checks, which are typically performed on a preliminary set of candidate results or by customized search indexes.




The quality of the retrieved images is typically evaluated by means of precision and recall measures. As in many other papers \cite{Philbin07,Jegou2009,Perronnin2010,Jegou2012}, we combined this information by means of the mean Average Precision (mAP), which represents the area below the precision and recall curve.

For evaluating the effectiveness of the recognition, which is basically a classification task, we use the micro-averaged \emph{accuracy} and macro-averaged $F_1$
(i.e., the harmonic mean of \emph{precision} and \emph{recall}).
Macro-averaged scores are calculated by first evaluating each measure for each category and then taking the average of these values.
Note that for a recognition task (i.e., single label classification), micro-averaged \emph{accuracy} is defined as the number of documents correctly classified divided by the total number of documents of the same label in the \emph{test set} and it is equivalent to the micro-averaged \emph{precision}, \emph{recall} and $F_1$ scores.

\label{sec:datasets}
For evaluating the performance of the various reduction techniques approaches, we make use of three datasets.
The first is the largely used Oxford Building datasets that was presented in \cite{Philbin07} and in many other papers. The dataset consists of 5,062 images of 55 buildings in Oxford. The ground truth consists of 55 queries and related sets of results divided in best, correct, ambiguous and not relevant. The dataset is intended for evaluating the effectiveness of a content based image retrieval systems that is expected to put the images related to the very same building at the top of the results list. In fact, the measure of performance used for the evaluation is the mean Average Precision (mAP).
The authors of the datasets also made available the words assigned to the images using the BoF approach. The vocabulary used has one million words.

We decided to use a second dataset to better evaluate the performance of a systems intended for recognizing the landmark in photos. In this scenario it is not important to retrieve most of the related images in the dataset but to correctly classify the image.
The Pisa dataset consists of 1,227 photos of 12 landmarks located in Pisa (also used in \cite{AmatoICART2011} and \cite{AmatoSISAP2011}).
The photos were crawled from Flickr.
The dataset is divided in a \emph{training set} ($Tr$) consisting
of 226 photos (20\% of the dataset) and a \emph{test
set} ($Te$) consisting of 921 photos (80\% of the
dataset).
The size of the vocabulary used for the experiments with the BoF approach is 10k.
In this context the performance measures used are typically accuracy, precision, recall and micro and macro-averaged $F_1$.

Finally, a larger dataset of about 400k images from the professional Alinari\footnote{http://www.alinari.it} archive was used for efficiency evaluation.
All the images were resized to have a maximum between width and height equal to 500 pixels before the feature extraction process.

\begin{figure}
\centering
\includegraphics[trim=10mm 0mm 0mm 0mm, width=0.5\textwidth]{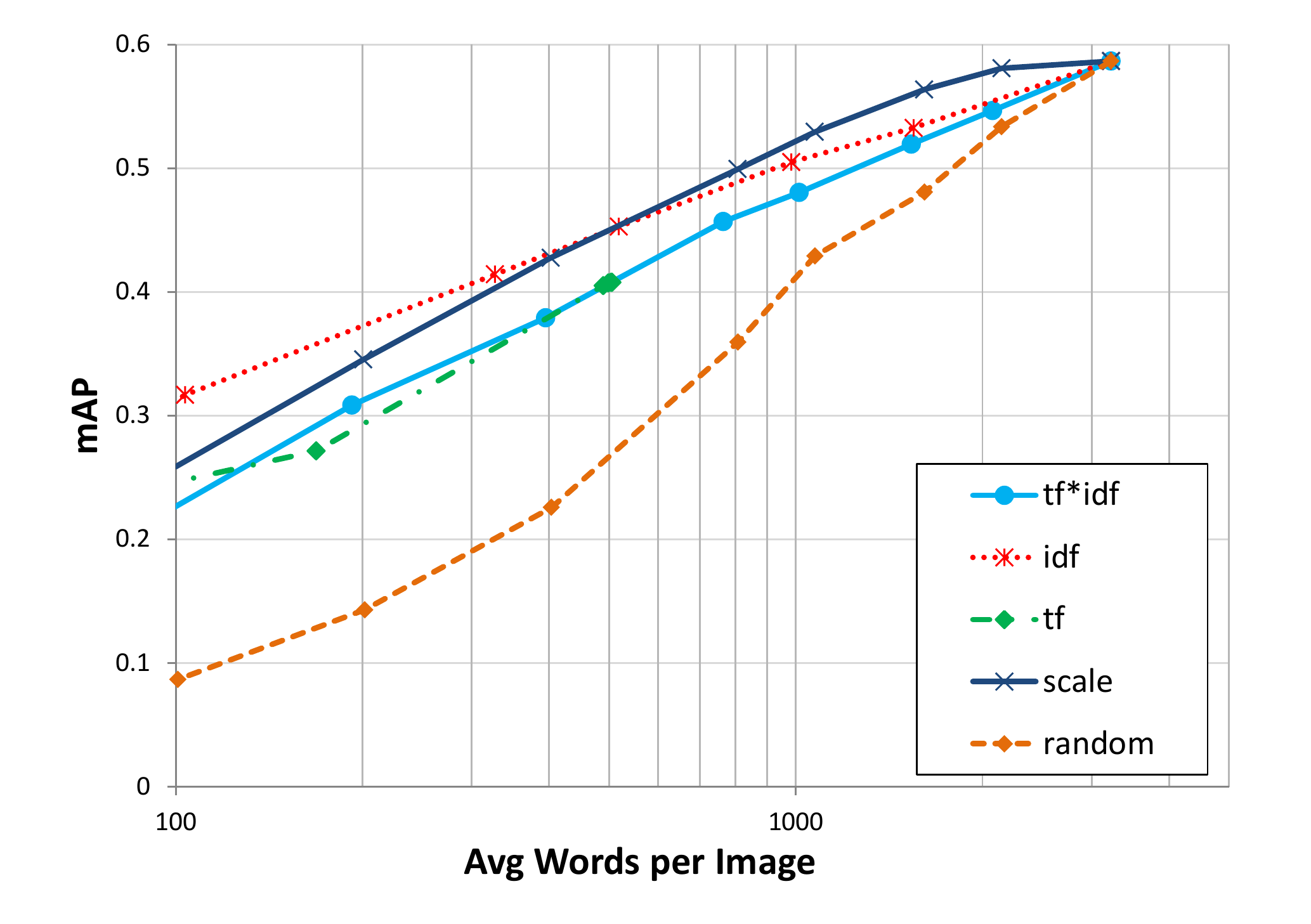}
\caption{Mean average precision of the various selection criteria obtained on the Oxford Buildings 5k dataset.}
\label{fig:map}
\end{figure}

\begin{figure}
\centering
\includegraphics[trim=10mm 0mm 0mm 0mm, width=0.5\textwidth]{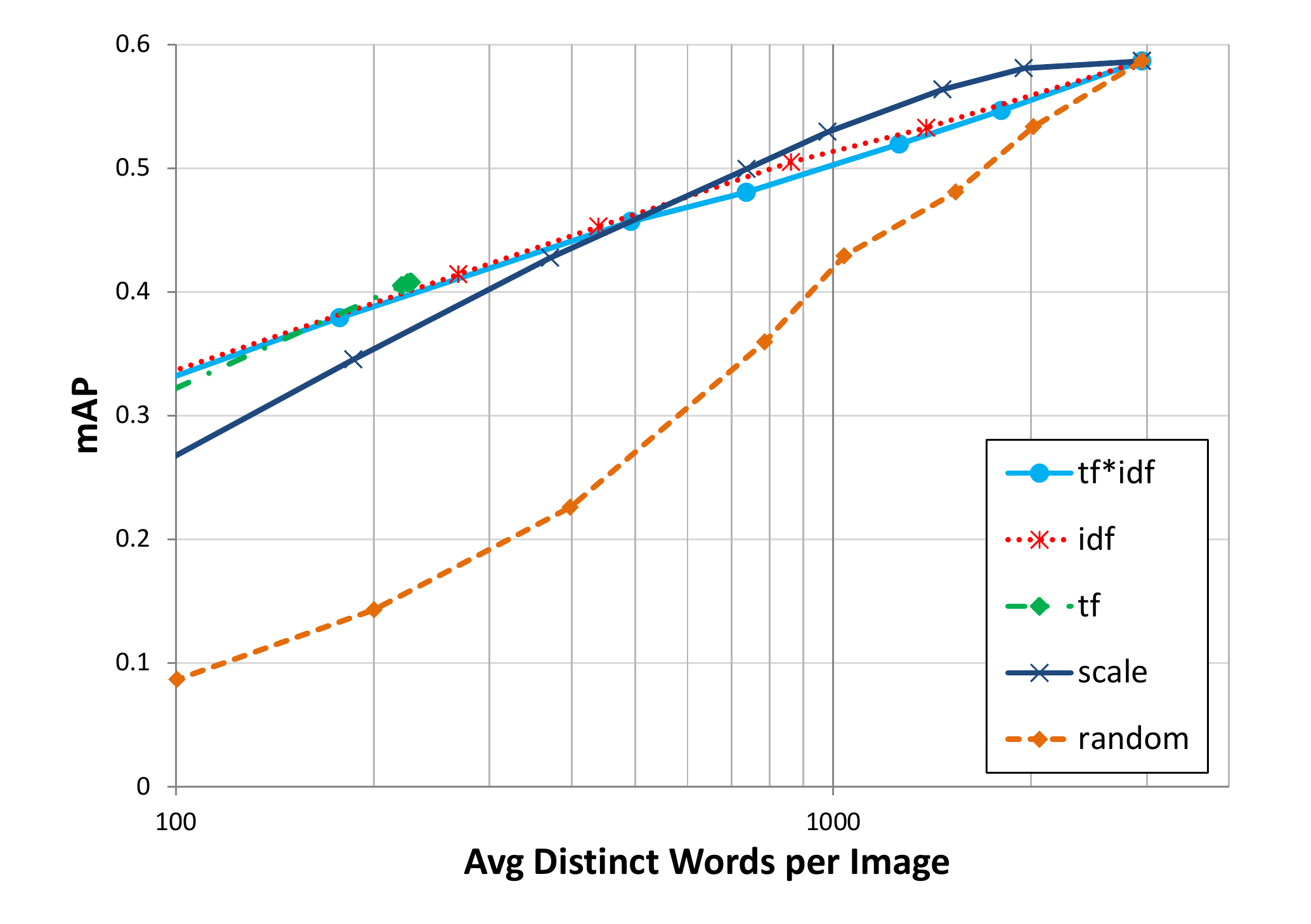}
\caption{Mean average precision of the various selection criteria obtained on the Oxford Buildings 5k dataset.}
\label{fig:map_dis}
\end{figure}

\subsection{Evaluation}

We first report the results obtained in a content based image retrieval scenario using the Oxford Building dataset using the ground truth given by the authors \cite{Philbin07}.
In Figure \ref{fig:map} we report the mAP obtained.
On the x-axis we reported the average words per image obtained after after the reduction. Note that the x-axis is logarithmic.
We first note that all the reduction techniques significantly outperform naive \emph{random} approach and that both the \emph{idf} and \emph{scale} approaches are able to achieve very good mAP results (about 0.5) while reducing the average number of words per image from 3,200 to 800.
Thus, just taking the 25\% of the most relevant words, we achieve the 80\% of the effectiveness.
The comparison between the \emph{idf} and \emph{scale} approaches reveals that \emph{scale} is preferable for reduction up to 500 words per image.
Please note that it is almost impossible to only slightly reduce the number of words with the \emph{tf} approach because there is a large number of words (about 75\%) per image that have just one occurrence. Using the \emph{tf} approach they have the same quality score and can be only filtered as a whole.

While the average number of words is useful to describe the length
of the image description, it is actually
the number of distinct words per image that have more impact on the efficiency
of searching using inverted index. Thus, in Figure \ref{fig:map_dis}, we report mAP with respect to the average number of distinct words. In this case the results obtained by \emph{tf*idf} and \emph{tf} are very similar to the ones obtained by \emph{idf}. In fact, considering \emph{tf} in the reduction results in a smaller number of average distinct words per image for the same vales of average number of words.

\begin{figure}
\centering
\includegraphics[trim=10mm 0mm 0mm 0mm, width=0.5\textwidth]{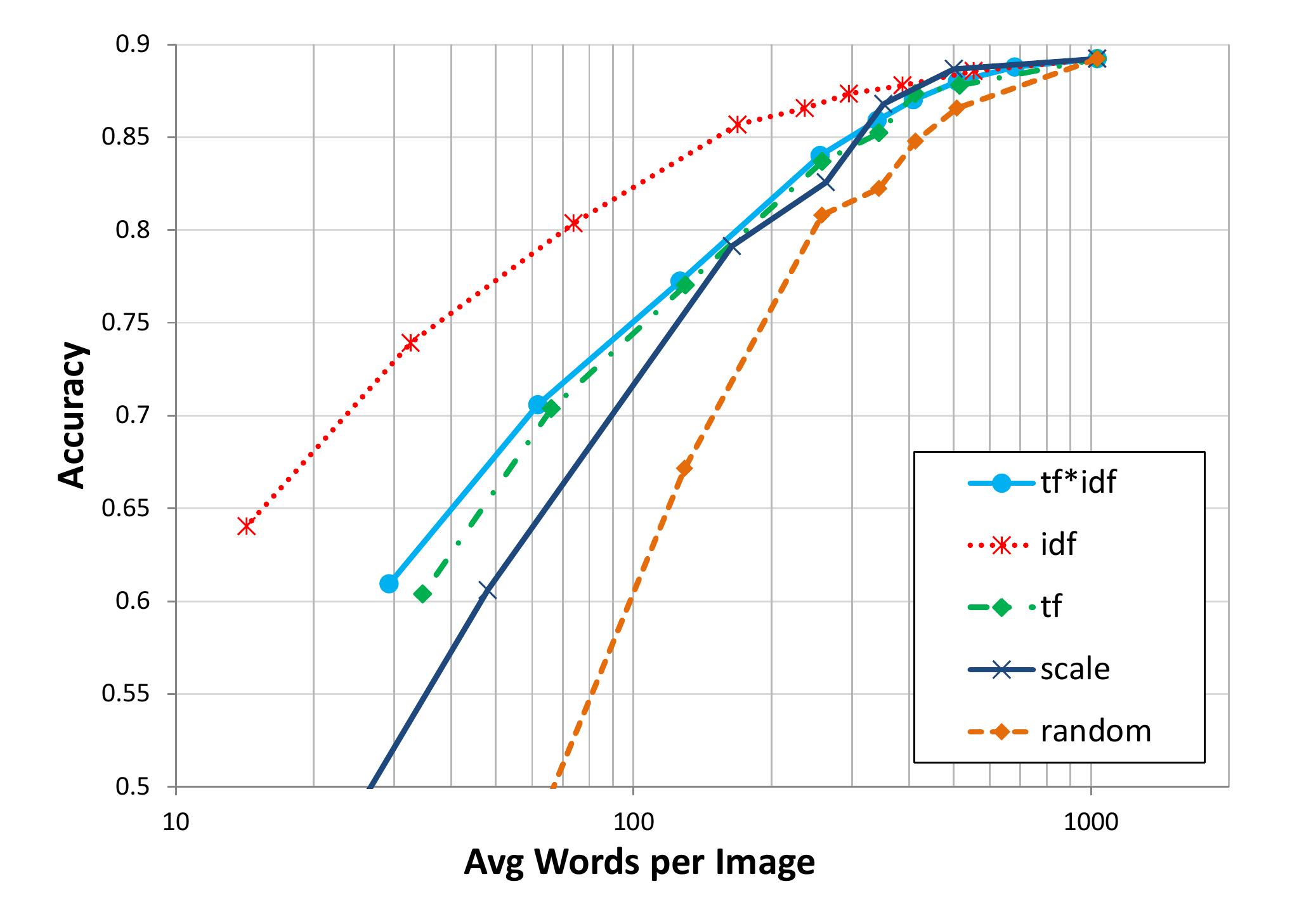}
\caption{Accuracy of the various selection criteria with respect to the average number of words per image on the Pisa dataset.}
\label{fig:acc}
\end{figure}

A second set of experiments was conducted on a landmark recognition task using the Pisa dataset (see Section \ref{sec:datasets}). For this dataset we used a smaller vocabulary of 10k words and features were extracted from a lower size images (maximum 512 pixels per side).
Figure \ref{fig:acc} reports the accuracy obtained by the various
approaches. On the x-axis we reported the average words per image
obtained after the reduction.
All the approaches, as expected, significantly outperform
the \emph{random} selection used as a baseline.
The best results are obtained by the \emph{idf} approach.
It is also interesting to notice that the \emph{scale}
approach performs very well for reduction up to 25\%.

\begin{figure}
\centering
\includegraphics[trim=10mm 0mm 0mm 0mm, width=0.5\textwidth]{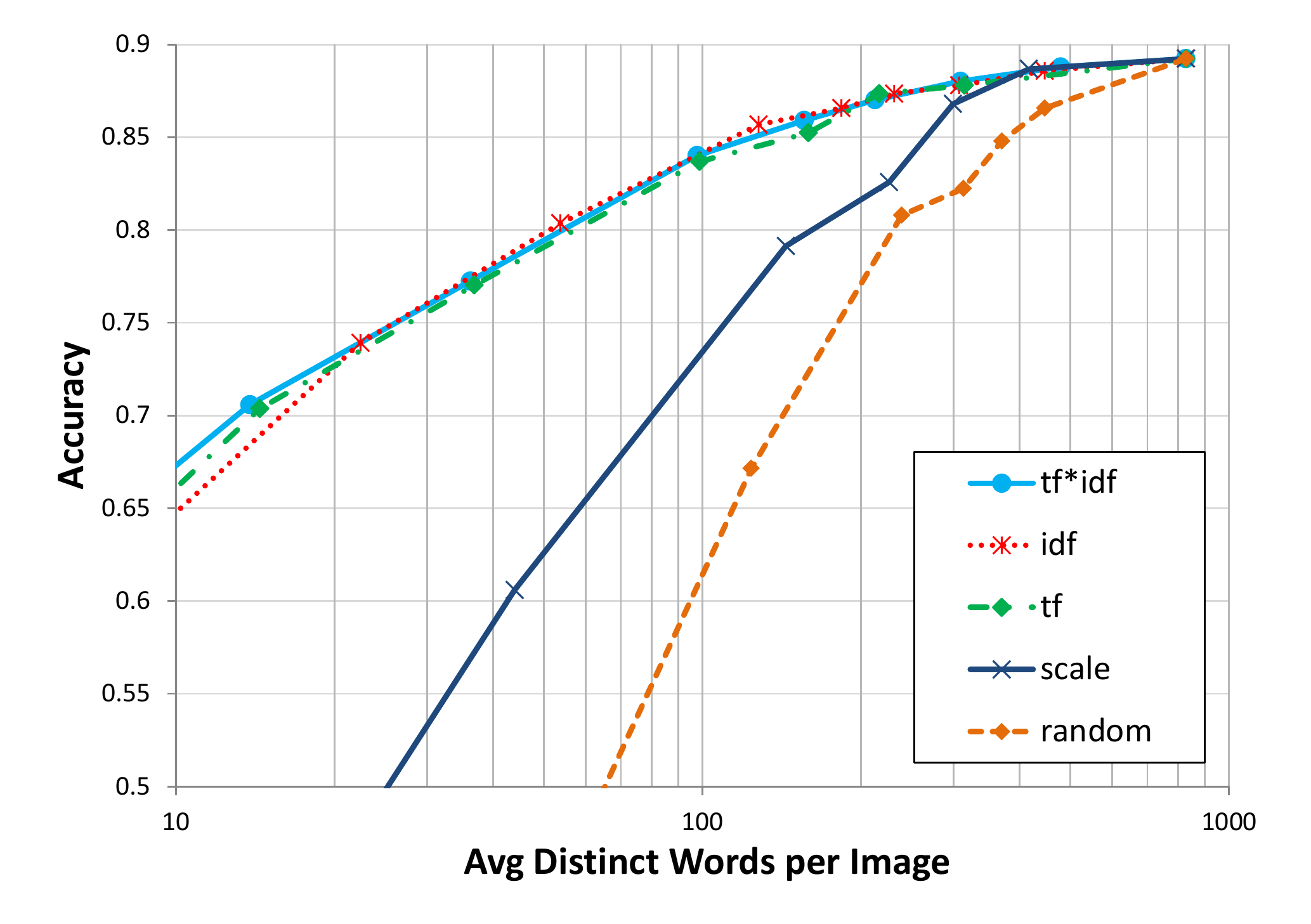}
\caption{Accuracy of the various selection criteria with respect to the average number of distinct words per image on the Pisa dataset.}
\label{fig:acc_dis}
\end{figure}

In Figure \ref{fig:acc_dis} we report the \emph{accuracy} obtained with
respect to the average number of distinct words per image by the
various reduction approaches. The results significantly differ
from the previous ones. In particular, the \emph{tf*idf} and
\emph{tf} approaches exhibit better results.
It is worth to say that while the \emph{tf*idf} approach relies on information about the training set for evaluating the \emph{idf} values, the \emph{tf} approach performs almost as better as \emph{tf*idf} can be applied
not considering the dataset but only the image for which the words
have to be reduced.
The \emph{scale} approach exhibits good
results for average number of distinct words down to 300 (i.e., a
25\% reduction).
Note that the \emph{scale} technique can be
applied before the words assignment thus reducing not only the search
cost but also the cost for the words assignment.

\begin{figure}
\centering
\includegraphics[trim=10mm 0mm 0mm 0mm, width=0.5\textwidth]{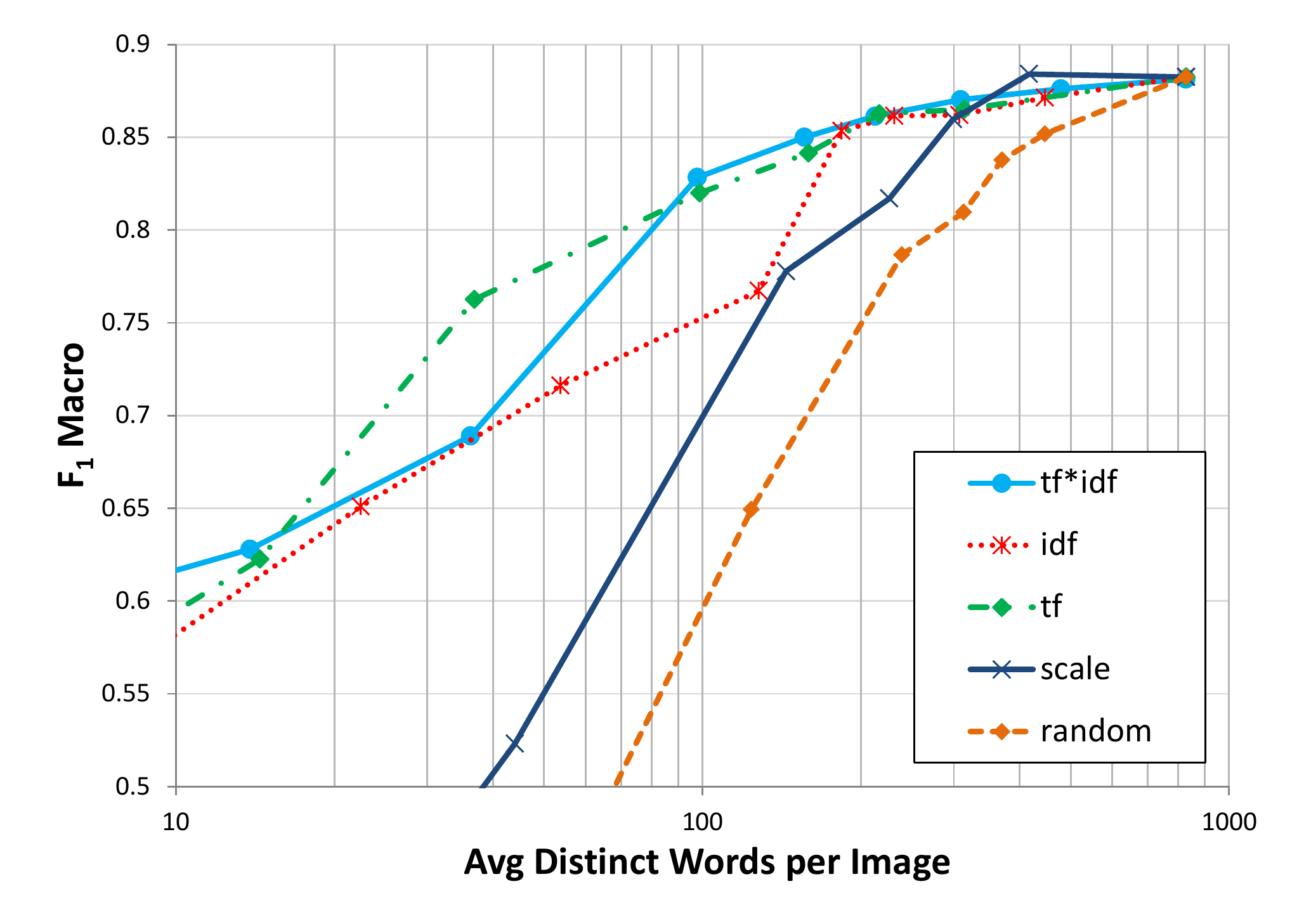}
\caption{Macro-averaged $F_1$ of the various selection criteria with respect to the average number of distinct words per image on the Pisa dataset.}
\label{fig:Mf1_dis}
\end{figure}

The \emph{accuracy} measure captures the overall effectiveness of
the algorithm with respect to the expected distribution of query
between the classes. In order to also evaluate the effectiveness
across the various classes (e.g., landmarks) we use the macro-averaged
$F_1$ measure. Macro-averaged values are calculated by first averaging
the measures obtained for each category.
In Figure \ref{fig:Mf1_dis}
we report the $F_1$ obtained by the various approaches in
terms of the average number of distinct words per image. The most
important differences between these results and the one obtained
considering \emph{accuracy} are related to the \emph{idf} and
\emph{Scale} approaches. While the \emph{scale} approach reveals
better performances for small reduction even increasing the
overall efficacy, the \emph{idf} results becomes worse than both
the \emph{tf} and \emph{idf} ones. The intuition is that
\emph{idf} relies on information related to the dataset and thus
is influenced by the different number of training images per
class. On the other hand, the \emph{scale} approach is independent
from the dataset, given that it does not rely on the words, thus
not even on the vocabulary.

\begin{figure}
\centering
\includegraphics[trim=10mm 0mm 0mm 0mm, width=0.5\textwidth]{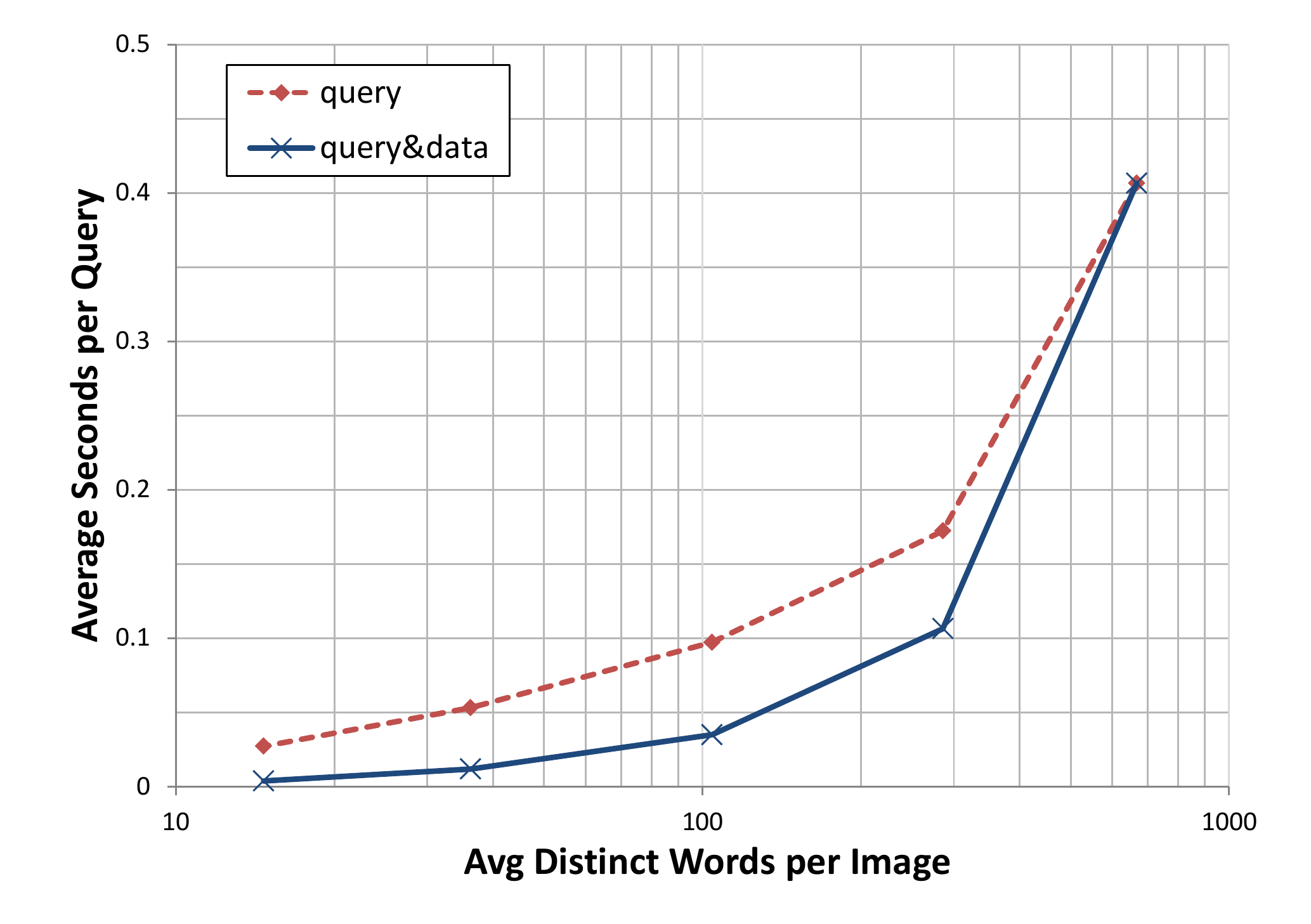}
\caption{Average search time with respect to the average number of distinct words per image obtained reducing the visual words on the query and on query and dataset}
\label{fig:eff}
\end{figure}

In Figure \ref{fig:eff} we report the average query execution time obtained on the 400k image dataset with respect to the average distinct words. Results are shown for reducing the visual words on query only and on query and dataset. While results are shown for the \emph{tf*idf} approach, similar performance are achieved with the other approaches. In fact, for efficiency, it is actually important only the average number of distinct words.
The results reveal, as expected, that high efficiency gains can be obtained reducing the number of distinct visual words. Note that the $x$-axis is logarithmic.

\section{\uppercase{Conclusion}}
\label{sec:conclusions}
\noindent In this work, we have investigated visual words reduction approaches in order to improve efficiency of the BoF approach minimizing the lost in effectiveness.
The gain in efficiency was tested on a similarity search scenario of about 400k images, while effectiveness was tested on two smaller datasets intended for content based image retrieval and landmark recognition.

We proposed methods that can be applied before the visual words have been assigned and also methods based on statistics of the usage of visual words in images (\emph{tf}), across the database (\emph{idf}), and on the \emph{tf*idf} combination.

In the content based image retrieval scenario the \emph{scale} approach performed best and even better than using all the words. However, for reduction over an order of magnitude effectiveness significantly decrease.
In the landmark recognition task, the most interesting results were obtained considering the macro-averaged $F_1$ effectiveness measure with respect to the average number of distinct words per image. The \emph{tf*idf} obtained the best results, but it is interesting to see that the \emph{tf} approach, which does not rely on dataset information, obtained very similar results. It is worth to note that the recognition task is more robust than the retrieval to words reduction.
Moreover, for small local features reductions \emph{scale} was the overall best.

We plan to define new approaches and compare with the ones proposed in this work on larger dataset in the near future.


%
%

\bibliographystyle{apalike}
{\small \bibliography{bib}}
%
%
\end{document}